\documentclass[nohyperref]{article}
\usepackage{fullpage}
\usepackage{minitoc}
\usepackage{url}
\usepackage{graphicx}
\usepackage{wrapfig}
\usepackage{amsmath}
\usepackage{amssymb}
\usepackage{microtype}
\usepackage{graphicx}

\usepackage{subcaption}
\usepackage{booktabs}
\usepackage{multirow}
\usepackage{wrapfig}
\usepackage{inputenc}
\usepackage{enumitem}
\usepackage{pdflscape}
\usepackage{rotating}
\usepackage{listings}
\usepackage{adjustbox}
\usepackage{comment}
\usepackage{pifont}
\usepackage{booktabs}
\usepackage{sectsty}

\sectionfont{\fontsize{13}{11}\selectfont}
\usepackage{multirow}
\usepackage{color, colortbl}
\usepackage[boxed]{algorithm2e}
\usepackage[authoryear,round]{natbib}
\usepackage[accsupp]{axessibility}  
\usepackage[most]{tcolorbox}
\usepackage[fixed]{fontawesome5}
\usepackage{authblk}

\title{\vspace{-1.1cm} Solving Spatial Supersensing Without Spatial Supersensing}

\definecolor{cvprblue}{rgb}{0.21,0.49,0.74}
\usepackage[pagebackref,breaklinks,colorlinks,citecolor=cvprblue]{hyperref}

\usepackage{cleveref}[2012/02/15]

\definecolor{scholarblue}{rgb}{0.21,0.49,0.74}
\definecolor{bluelink}{RGB}{0,113,188}
\definecolor{greenlink}{RGB}{0,188,113}
\definecolor{anthro}{RGB}{246,244,238}
\hypersetup{
    colorlinks=true,%
    citecolor=scholarblue,%
    filecolor=red,%
    linkcolor=red!93!black,%
    urlcolor=bluelink
}
\newcommand{\finding}[2]{
    \begin{tcolorbox}[
        colback=white!50!anthro,     
        colframe=anthro!60!black,     
        arc=5pt,                    
        boxsep=5pt,                 
        left=10pt,                  
        right=10pt,                 
        top=2pt,                    
        bottom=2pt,                 
        boxrule=0.8pt,              
        drop shadow=gray!50!white,  
        enhanced jigsaw             
    ]
    #2
    \end{tcolorbox}
}

\crefformat{footnote}{#2\footnotemark[#1]#3}
\Crefname{table}{Table}{Tables}
\crefname{table}{Tab.}{Tabs.}
\Crefname{figure}{Figure}{Figure}
\crefname{figure}{Fig.}{Figs.}
\Crefname{appendix}{Appendix}{Appendix}
\crefname{appendix}{Appx.}{Apps.}
\Crefname{algorithm}{Algorithm}{Algorithm}
\crefname{algorithm}{Alg.}{Algs.}
\Crefname{section}{Section}{Section}
\crefname{section}{Sec.}{Secs.}

\date{}
\makeatletter
\renewcommand\AB@affilsepx{, \protect\Affilfont}
\makeatother
\author{Vishaal Udandarao$^1$ \quad Shyamgopal Karthik$^1$ \quad Surabhi S. Nath$^2$ \quad Andreas Hochlehnert$^1$ \quad Matthias Bethge$^1$ \quad Ameya Prabhu$^1$\\\vspace{-0.2cm}{\normalsize $^1$Tübingen AI Center, University of Tübingen \quad $^2$Max Planck Institute for Biological Cybernetics}}

\begin{document}
\maketitle

\doparttoc 
\faketableofcontents 

\vspace{-1.2cm}
\begin{center}
    \begin{tabular}{c@{\hskip 19pt}c}

    \hspace*{1.4cm}\raisebox{-1.5pt}{\faGithub} \href{https://github.com/bethgelab/supersanity}{\fontsize{10pt}{0pt}\path{github.com/bethgelab/supersanity}} & \\
\end{tabular}
\end{center}

\vspace{-0.25cm}
\begin{abstract}
\noindent Cambrian-S aims to take the first steps towards improving video ``world models'' with spatial supersensing by introducing (i) two benchmarks, VSI-Super-Recall (VSR) and VSI-Super-Counting (VSC), and (ii) bespoke \textit{predictive sensing} inference strategies tailored to each benchmark. 
In this work, we conduct a critical analysis of Cambrian-S across both these fronts.
First, we introduce a simple baseline, \textit{NoSense}, which discards almost all temporal structure and uses only a bag-of-words SigLIP model, yet near-perfectly solves VSR, achieving $95\%$ accuracy even on 4-hour videos. This shows benchmarks like VSR can be nearly solved \textit{without} spatial cognition, world modeling or spatial supersensing. 
Second, we hypothesize that the tailored inference methods proposed by Cambrian-S likely exploit shortcut heuristics in the benchmark. We illustrate this with a simple sanity check on the VSC benchmark, called \textit{VSC-Repeat}: We concatenate each video with itself 1-5 times, which does not change the number of unique objects. However, this simple perturbation entirely collapses the mean relative accuracy of Cambrian-S from $42.0\%$ to $0\%$. A system that performs spatial supersensing and integrates information across experiences should recognize views of the same scene and keep object-count predictions unchanged; instead, Cambrian-S’ inference algorithm relies largely on a shortcut in the VSC benchmark that rooms are never revisited. Taken together, our findings suggest that (i) current VSI-Super benchmarks do not \textit{yet} reliably measure spatial supersensing, and (ii) predictive-sensing inference recipes used by Cambrian-S improve performance by inadvertently exploiting shortcuts rather than from robust spatial supersensing. We include the response from the Cambrian-S authors (in Appendix \ref{sec:response}) to provide a balanced perspective alongside our claims.
\end{abstract}

\begin{figure}[!ht]
    \centering
    \includegraphics[width=0.89\linewidth]{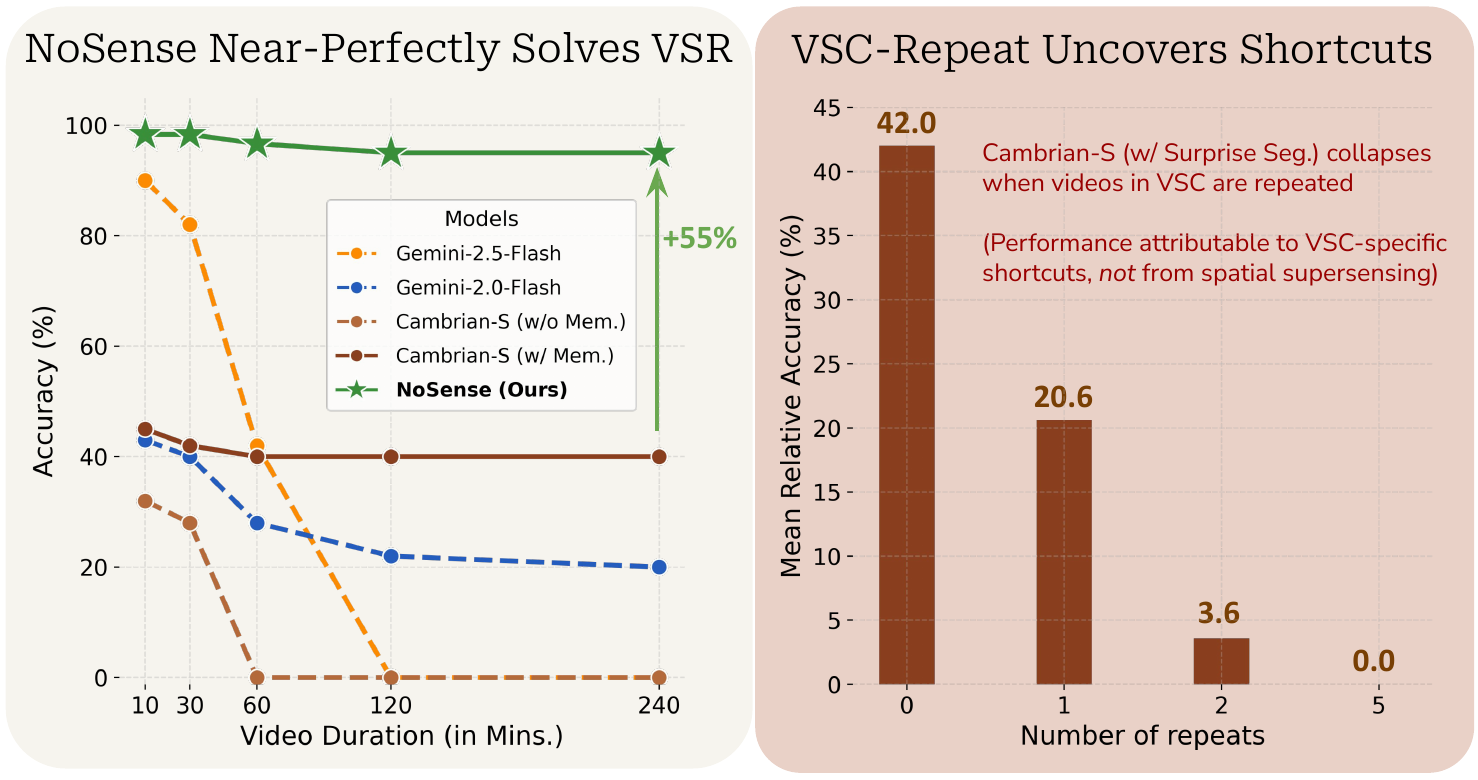}
    \vspace{-0.2cm}
        \caption{\textbf{(Left) \textit{NoSense} solves VSR without supersensing.} Our \textit{NoSense} baseline uses only a SigLIP model with independent frame-level processing -- no video model, LLM, long-term memory, or temporal reasoning -- yet almost perfectly solves VSR, showing that the VSR benchmark can be solved without spatial supersensing. \textbf{(Right) Cambrian-S exploits VSC-specific shortcuts.} For the VSC benchmark, we repeat each 10-min video 1–5 times; a supersensing model should output the same object counts, since the \textit{unique} object count stays the same. Instead, Cambrian-S' mean relative accuracy collapses from 42\% to 0\% after 5 repeats, indicating that its surprise-based segmentation inference method relies on VSC-specific shortcuts rather than genuine spatial cognition.}
    \label{fig:teaser}
    \vspace{-0.4cm}
\end{figure}

\section{Introduction}
Video understanding benchmarks should test whether models build coherent, predictive world models from long video streams. This requires accumulating visual evidence, maintaining state, and tracking entities over time---abilities that humans use naturally but current models largely lack. In practice, many benchmarks do not meet this goal. Prior work shows that models can often ignore most of the temporal structure in the video, rely on a single frame, or even use text-only inputs and still match or surpass state-of-the-art \textit{video} models~\citep{cambrian-s,tangemann2020measuring,buch2022revisiting,lei2023revealing,li2018resound,goyal2017making, brown2025benchmark}. Much of what we call \textit{video understanding} reduces to image captioning or language modeling with minimal temporal reasoning.\vspace{0.2cm}

\noindent This is known as shortcut learning~\citep{geirhos2020shortcut}. Benchmark designers aim for tasks that are hard to solve via shortcuts, so that the easiest way to improve performance is to develop the intended capability \citep{brown2025benchmark}. When this fails, the design can unintentionally nudge practitioners into baking those same shortcuts into their models \citep{hammoud2023rapid,dehghani2021benchmark}. Large models often similarly exploit spurious cues instead~\citep{geirhos2020shortcut}. A recent work, Cambrian-S~\citep{cambrian-s} proposes to address these issues.\vspace{0.2cm}

\noindent Cambrian-S argues that benchmarks should target specific capabilities: streaming event cognition, implicit 3D reasoning, and \textit{spatial supersensing} i.e., whether models form predictive world models from raw video. To resist brute-force context expansion, Cambrian-S introduces the VSI-Super benchmarks which need true spatial reasoning abilities. The tasks use long, multi-environment videos and questions designed to require accumulating evidence, tracking objects, and maintaining state over time. The promise is clear: strong performance on these benchmarks should require genuine spatial cognition and memory. This motivates our central questions about VSI-Super:\vspace{0.2cm}

\begin{enumerate}[leftmargin=*]
    \item Does success on \emph{VSI-Super-Recall} (VSR) truly require supersensing or even spatial cognition, or can a simple SigLIP model which discards almost all temporal structure still perfectly solve the benchmark?
    \item Does the Cambrian-S inference method perform the intended event-level reasoning and spatial supersensing, or does it rely on shortcuts specific to the benchmark like \emph{VSI-Super-Counting} (VSC)?
\end{enumerate}

\noindent To investigate these questions, we conduct two complementary analyses:
\begin{enumerate}[leftmargin=*]
    \item \textbf{VSR: A Simple Baseline.} We take SigLIP-2, select the four frames most relevant to the object mentioned in the question, and select the most correlated answer option. We call the resulting baseline \textit{NoSense}, short for \textit{No Supersensing}. NoSense uses no explicit 3D representation, does not track objects, and does not use language models. Despite this simplicity, NoSense nearly perfectly solves VSR, even on 4-hour videos, while discarding almost all temporal structure in the video. This suggests that VSI-Super-Recall, like many prior video benchmarks, primarily measures image-level semantic cues and object–context associations rather than spatial supersensing, and only needs a few informative frames rather than long-range temporal understanding.
    \item \textbf{VSC: A Sanity Check.} Does the inference method of Cambrian-S, designed for each benchmark, simply exploit shortcuts in the benchmark? We evaluate this by creating \textit{VSC-Repeat}: We take each video clip in the VSC benchmark and concatenate it with itself 1-5 times, creating a sequence in which every room and object is revisited repeatedly. A model performing genuine supersensing should recognize that these are repeated observations of the same scene; therefore, the predicted object counts should remain unchanged. However, repeating the video a few times reduces the mean relative accuracy of Cambrian-S from $42\%$ to $0\%$. This indicates the inference algorithm inadvertently exploits the benchmark shortcuts rather than improving supersensing.
\end{enumerate}

\noindent Our findings do not imply that there has been no progress in improving spatial supersensing. Gemini 2.5 Flash achieves around 41.5\% accuracy on 60-minute VSR videos, illustrating that substantial VSR performance can arise from generic long-context video training. While VSI-Super is a comprehensive, thoughtful effort, it still fails to reliably enforce the long-horizon spatial supersensing it aims to elicit. Cambrian‑S’ VSR inference pipeline and \textit{NoSense} share a similar high‑level pattern -- streaming frame encoding, a compact memory of salient frames, and query‑guided retrieval -- suggesting that a much simpler instantiation of this pattern already goes a long way on VSR. Our simple sanity check, \textit{VSC-Repeat} shows the degree to which VSC-tailored inference algorithms rely on shortcuts. Our two-fold analysis underscores the need for caution when interpreting performance on VSI-Super: rather than reflecting genuine spatial cognition, high scores may simply mean exploiting heuristic shortcuts. We further include a response from the authors (in Appendix \ref{sec:response}) to provide a balanced perspective alongside our claims. \vspace{0.2cm}

\noindent Overall, our tools -- \textit{NoSense} for VSR and \textit{VSC-Repeat} for VSC -- act as simple stress tests for benchmarks and training methods. We view this kind of meta-evaluation as analogous to ablation studies for models and believe it should be standard for benchmarks. For any proposed spatial supersensing task, we should ask: what is the simplest streaming baseline that exploits obvious regularities? Which invariances should hold, and do methods respect them? 
We hope our analysis fosters more rigorous testing and design of superspatial sensing in the future.

\section{Spatial Supersensing Benchmarks Don’t Need Spatial Supersensing }

\begin{figure}[h]
    \centering
    \vspace{-0.5cm}
\includegraphics[width=0.975\linewidth]{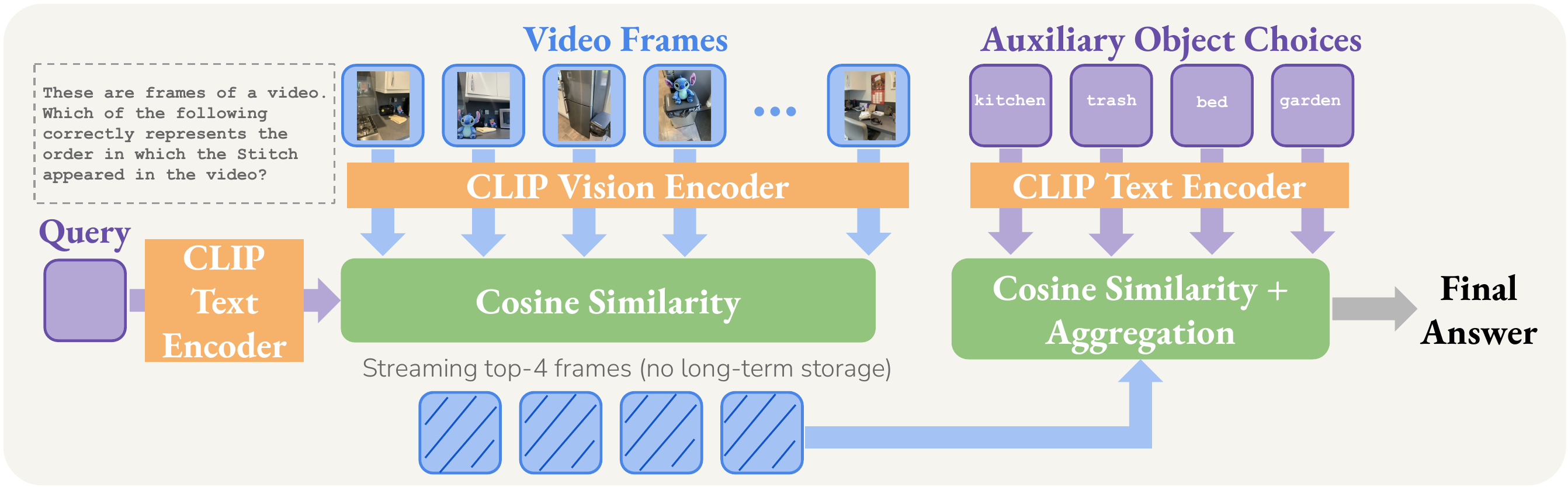}
    \vspace{-0.3cm}\caption{\textbf{\textit{NoSense} does no spatial supersensing.} Frames are encoded independently with SigLIP. We keep top-4 frames by cosine similarity to the object query and aggregate similarities to auxiliary objects from the MCQ options to select the answer. The pipeline is streaming, memory-efficient and uses only the relative order of the four most object‑relevant frames; it never reasons about motion, continuity, or long-range temporal patterns.}
    \vspace{-0.3cm}
    \label{fig:pipeline}
\end{figure}

\noindent In this section, we introduce \textit{NoSense}, a streaming, atemporal baseline for the VSR benchmark. We first state the task and method, then report our results compared to the previous SoTA, and finally present ablation studies.\vspace{0.2cm}

\noindent \textbf{Task.} VSR tests whether a system can recover the appearance order of an object of interest. Each video $\mathcal{V}$ contains four inserted frames in which the same object of interest (e.g., a teddy bear) is co-located with $4$  auxiliary objects or environments (e.g., bed, bathtub, sink, floor). The goal is to predict the correct temporal ordering of these four environments from videos lasting $10$ minutes to $4$ hours.\vspace{0.2cm}

\begin{algorithm}[b]
\DontPrintSemicolon
\caption{\textit{NoSense} for VSR: streaming CLIP/SigLIP with no temporal reasoning.}
\label{alg:nosense-compact}
\KwIn{Frames $\{x_t\}_{t=1}^N$ at 1 FPS; object $o$; auxiliaries $\{a_i\}_{i=1}^4$; answer options $\{\pi^{(k)}\}_{k=1}^4$}
\KwOut{$\hat{k}$}
Encode $o$ with a prompt ensemble to get normalized $O\in\mathbb{R}^d$\;
Encode each frame: $F_t=\mathrm{norm}(f_{\text{img}}(x_t))$\;
$s_t=\langle F_t,O\rangle$ for $t=1,\dots,N$; keep top-4 indices $(t_1<\dots<t_4)$ in buffer while streaming\;

Encode joint text for $(o,a_i)$ to get normalized $A_i$.\;
$r_{u,i}=\langle F_{t_u},A_i\rangle$ for $u,i\in\{1,\dots,4\}$\;
\For{$k\in\{1,\dots,4\}$}{ $\text{score}(k)=\sum_{u=1}^4 r_{u,\pi^{(k)}_u}$ }
$\hat{k}=\arg\max_k \text{score}(k)$\; 
\Return $\hat{k}$\;
\end{algorithm}

\noindent \textbf{NoSense baseline.} \textit{NoSense} is a simple streaming method that uses only a CLIP/SigLIP model~\citep{radford2021learning,zhai2023sigmoid}, illustrated in Figure \ref{fig:pipeline}. It assumes a video of $N$ frames sampled at 1 FPS, an object of interest $o$, four auxiliaries $\{a_i\}_{i=1}^4$, and four answer options $\{\pi^{(k)}\}_{k=1}^4$ (each a permutation of the auxiliaries). NoSense performs the following steps: (i) it processes frames independently (discards nearly all temporal information), (ii) it keeps only the top four frames most similar to the query object in a buffer (no long-term memory), and (iii) it scores answer options by aggregating cosine similarities to auxiliary object prompts (no LLM). It requires no spatial supersensing, spatial cognition, or world modeling. It uses the same information and procedural constraints: query-aware, frame-level retrieval paradigm employed in Cambrian-S’ own VSR inference pipeline. Thus, NoSense should be interpreted as a valid solution, not as a misuse of the benchmark. The full algorithm is in Algorithm \ref{alg:nosense-compact}. \vspace{0.2cm}

\noindent \textbf{Experimental details.} The primary and only main design choice for \textit{NoSense} is the contrastive VLM used for text and image embeddings. We evaluate four models: (i) OpenAI-CLIP-L/14, (ii) SigLIP2-B/16, (iii) SigLIP2-So400m-384, and (iv) SigLIP2-So400m-512. Our best results are obtained with SigLIP2-So400m-512.\vspace{0.2cm}

\noindent \textbf{NoSense (almost) perfectly solves the VSR benchmark.} As shown in Figure \ref{fig:teaser}, \textit{NoSense} nearly solves all five test splits, from 10-minute to 4-hour videos. It achieves $98.3\%$ on the 10-minute split and maintains $\approx95\%$ on the 2- and 4-hour splits, outperforming Cambrian-S by $55\%$ in absolute performance. These results show that VSR does not require explicit world modeling or long-horizon reasoning; a simple contrastive-VLM baseline which exploits shortcuts in the benchmark by adopting a design similar to the VSR-specific inference algorithm suffices.\vspace{0.2cm}

\finding{}{\faBookmark~~NoSense shows that the VSI-Super-Recall (VSR) benchmark overwhelmingly focuses on semantic perception while neglecting the more advanced spatial and temporal reasoning required for supersensing.}

\noindent \textbf{Why does NoSense work so well?} \textit{NoSense} succeeds without extra supervision or training because the benchmark construction tightly aligns with contrastive retrieval pipeline used by the inference algorithm of Cambrian-S. We show that (i) \textit{NoSense} uses no more inductive bias than Cambrian-S for this task, and (ii) relaxing our text-embedding design still yields strong performance.\vspace{0.2cm}

\begin{figure}[!t]
    \centering
    \vspace{-0.4cm}
    \includegraphics[width=0.9\linewidth]{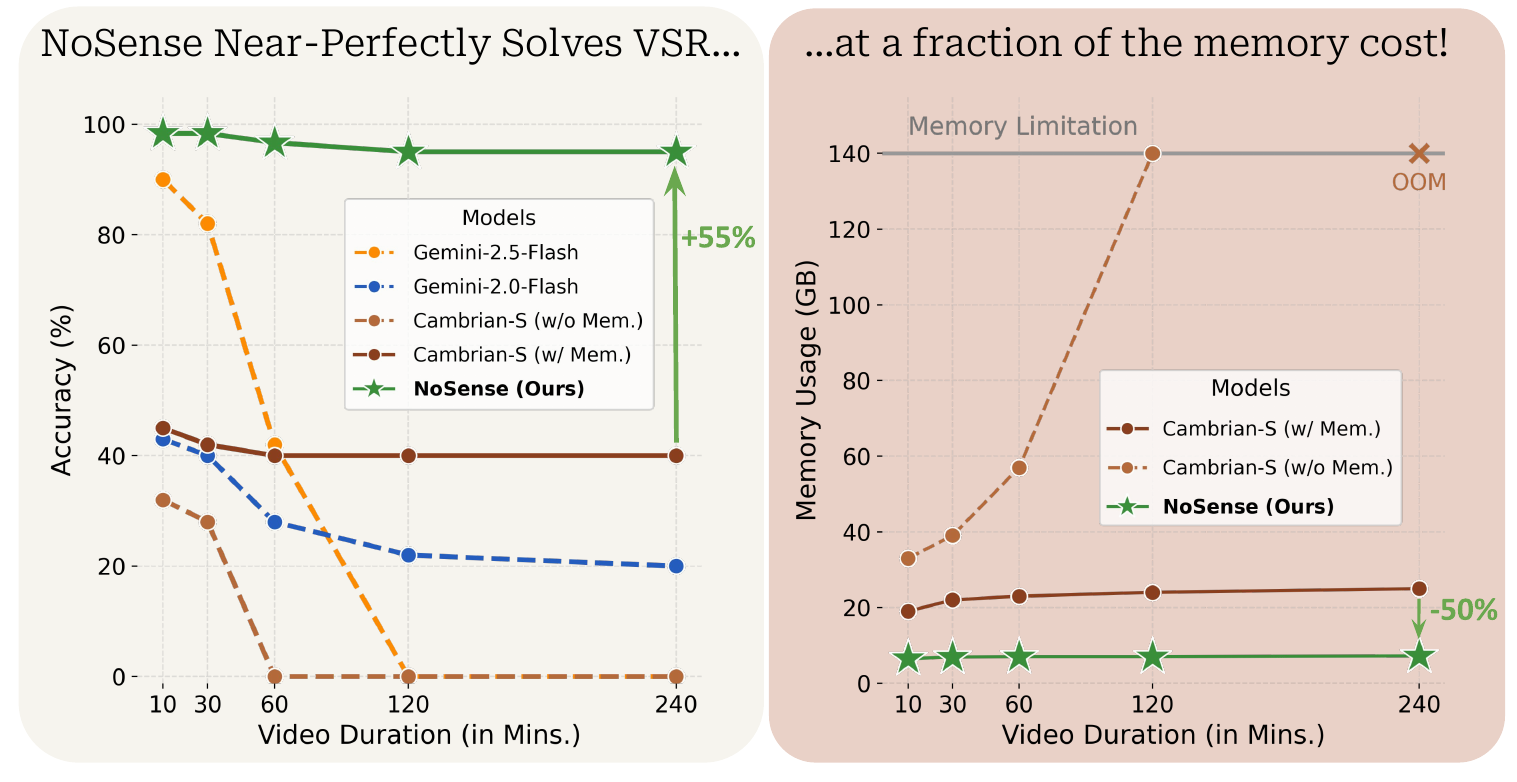}
    \vspace{-0.3cm}
        \caption{\textbf{\textit{NoSense} solves VSR with no supersensing.} \textit{NoSense} uses only a SigLIP image encoder with independent frame-level processing—no video model, LLM, memory, or temporal reasoning. Yet, NoSense nearly perfectly solves VSR \textit{(left)}, while using a fraction of the GPU memory of previous methods \textit{(right)}. This clearly shows that VSR can be solved without explicit 3D state, object tracking, or long‑horizon temporal reasoning.}
        \vspace{-0.3cm}
    \label{fig:nosense-vsr}
\end{figure}

\begin{table}[t]
\centering
\vspace{-0.3cm}
\caption{\textbf{Ablation Results}. Across all variants, NoSense remains far ahead of Cambrian-S on VSR. (\textit{left}) Even when we (1) use the raw question without extracting the target object, or (2) drop prompt ensembling and use a single basic template. (\textit{right}) Likewise, swapping the SigLIP-2 encoder for weaker CLIP/SigLIP variants still leaves us well above Cambrian-S on every split. This robustness suggests that near-perfect performance on VSR is not a fragile artifact of prompt engineering or a particular encoder.}
\begin{minipage}{0.45\textwidth}
\vspace{-0.35cm}
    \resizebox{\textwidth}{!}{%
    \begin{tabular}{lccc}
    \toprule
    \textbf{Method} & \textbf{10 mins} & \textbf{30 mins} & \textbf{60 mins} \\
    \midrule
    \textcolor{gray}{Cambrian-S (baseline)} & \textcolor{gray}{45.0\%}  & \textcolor{gray}{41.7\%}  & \textcolor{gray}{40.0\%}  \\
    \textit{NoSense} (Ours, Best) & 98.3\%  & 98.3\%  & 96.7\%  \\
    \midrule
    \multicolumn{4}{l}{\textbf{Ablation 1: Use Raw Questions}} \\
    NoSense (Raw) & 90.0\% & 81.7\% & 75.0\% \\
    \midrule
    \multicolumn{4}{l}{\textbf{Ablation 2: Use Basic Prompt}} \\
    NoSense (Basic Prompt) & 98.3\% & 98.3\% & 93.3\% \\
    \bottomrule
    \end{tabular}%
    }
\end{minipage}
\begin{minipage}{0.275\textwidth}
    \vspace{-0.5cm}
    \includegraphics[width=\textwidth]{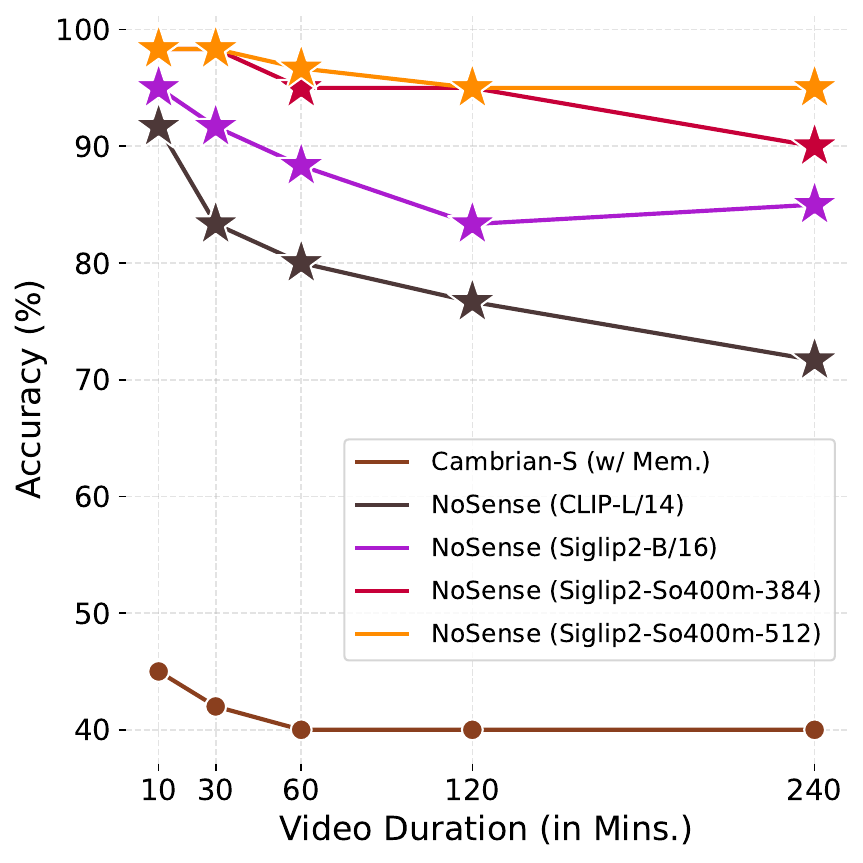}
    \vspace{-0.5cm}
\end{minipage}
\vspace{-0.4cm}
\label{tab:ablation}
\end{table}

\noindent \textit{(i) Cambrian-S and \textit{NoSense} share the same functional components.} Both methods use: (a) a vision encoder for the stream, (b) a memory with consolidation (top-$k$ frames for \textit{NoSense}), and (c) query-guided retrieval mechanism. \textit{NoSense} implements these aspects with a contrastive VLM rather than a multimodal LLM, and removes components such as long-term memory that are necessary for spatial supersensing but not for this benchmark. The fact that a simpler instantiation near-perfectly solves VSR suggests that it can be solved without additional inductive biases.
Given this similarity, it is plausible that Cambrian‑S’ VSR gains also rely heavily on the same basic retrieval‑style shortcut -- identifying a few discriminative frames for the target object -- rather than by better spatial supersensing.\vspace{0.2cm}

\finding{}{\faBookmark~~Cambrian-S uses two different inference pipelines across the two VSI-Super benchmarks (VSR and VSC), which may rely on shortcuts to improve performance without improving spatial supersensing.}

\vspace{-0.25cm}
\paragraph{Why was this missed?} Cambrian-S makes a careful effort to test shortcuts and strong baselines. However:
\begin{itemize}[leftmargin=*]
    \item Cambrian-S uncovers flaws with existing benchmarks, these were done by measuring the performance of language-only, single-frame and caption-only baselines. Single-frame baselines cannot directly work on VSR, because it requires at least four frames. Caption-based pipelines \emph{could} work, but the implementation in Cambrian-S summarizes only 32 frames (as in short-video setups like LLaVA-OneVision~\citep{llavaonevision}), which dramatically undersamples long videos. Captioning at $\sim$1 FPS might suffice.
    \item Cambrian-S evaluates long-video methods such as MovieChat~\citep{moviechat} and Flash-VStream~\citep{flashvstream} specifically for these tasks. However, they use generic memory mechanisms that are not designed to exploit VSR’s construction as directly as the inference pipeline of Cambrian-S.
\end{itemize}

\noindent \textit{(ii) Relaxing our text-embedding design still yields strong performance.} Until now, for extracting our text embeddings, we leveraged a prompt ensemble in our experiments as this has been shown to reliably improve embedding quality~\citep{radford2021learning}. We now study two simplified variants: (1) using the raw question text and skipping explicit extraction of the object from the question, and (2) removing prompt ensembling and using a single basic template for all text encodings. Additionally, we study robustness across 4 different CLIP/SigLIP models.\vspace{0.2cm}

\noindent \textbf{Ablation results.} Table \ref{tab:ablation} summarizes our ablations. We show results of using the raw question (ablation 1) in Table \ref{tab:ablation} (left, middle). We observe that performance drops significantly (8-20\%) compared to our best NoSense but remains far above Cambrian-S by 35-40\%. The target objects in VSR are distinctive enough that a CLIP-style text encoder produces discriminative signal directly from the raw query due to its bag-of-words nature~\citep{yuksekgonul2022and}, making explicit extraction optional. We next show results of removing prompt ensembling (ablation 2) in Table \ref{tab:ablation} (left, bottom). We observe that accuracy decreases slightly but performance is still extremely competitive (more than 50\% gain over Cambrian-S), indicating that prompt ensembling helps slightly but our results are largely independent of this aspect. Lastly, we test robustness across CLIP/SigLIP variants in Table \ref{tab:ablation} (right). Even when varying contrastive VLMs both in model family and size, \textit{NoSense} remains far ahead of Cambrian-S. This robustness suggests that relatively small contrastive VLMs already provide features sufficient to almost perfectly solve VSR. \vspace{0.2cm}

\noindent \textbf{Conclusion.} Our results indicate that, despite careful design, VSR can be solved primarily through semantic cues from a handful of informative frames, with little need for explicit 3D reasoning, state maintenance, or long-term memory. This gap between the intended and actual difficulty of VSR is closely related to shortcuts introduced in the benchmark construction. Our NoSense baseline effectively isolates this shortcut. Cambrian‑S’ VSR inference pipeline implements the same functional pattern -- vision encoding, a compact memory of salient frames, and query-guided retrieval -- but with a multimodal LLM. The fact that our simple baseline almost perfectly solves VSR suggests that a substantial portion of Cambrian-S’ VSR performance may come from a similar retrieval-style shortcut, rather than uniquely from the additional machinery intended to support spatial supersensing or world modeling. In the next section, we rigorously test this hypothesis on the VSC benchmark.

\section{Spatial Supersensing Methods Don't Do Spatial Supersensing}

\begin{figure}[!h]
    \centering
    \vspace{-0.4cm}
    \includegraphics[width=0.95\linewidth]{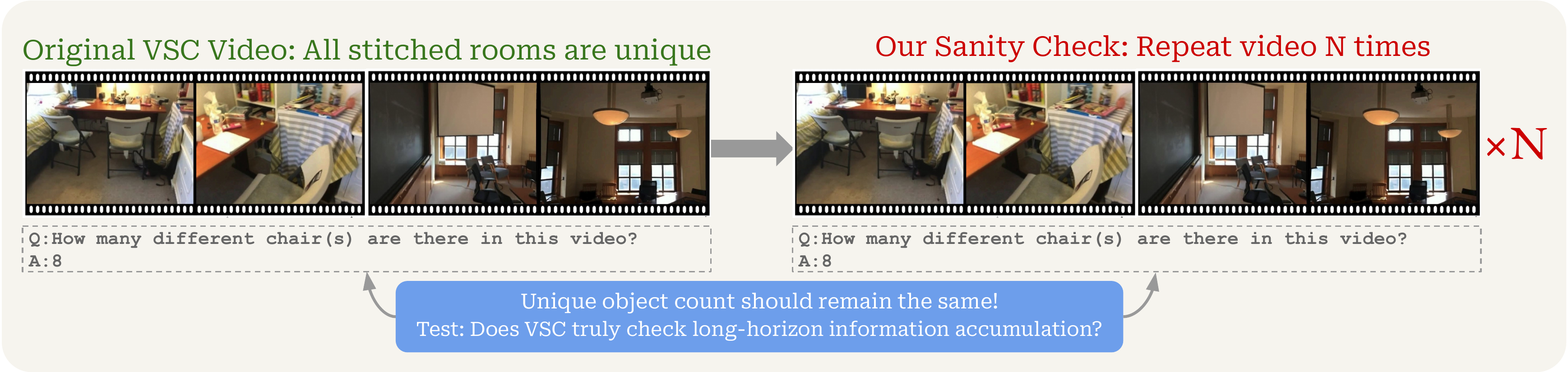}
    \vspace{-0.1cm}
    \caption{\textbf{Repeating VSC videos exposes a counting shortcut.} We propose a simple sanity check for VSC, called \emph{VSC-Repeat}. We  concatenate each VSC video (from the 10-min split) with itself $1-5$ times. Since no new scene is introduced, the ground-truth number of unique objects remains unchanged. This sanity check can help test if models indeed hold long-term global state or rather exploit simple segmentation-based shortcuts.}
    \vspace{-0.2cm}
    \label{fig:vscrepeatedbench}
\end{figure}

\noindent Previously, we showed that a simple frame-level baseline nearly saturates VSR without any spatial supersensing. This led to a hypothesis that Cambrian-S' benchmark-tailored inference methods might be improving performance by exploiting benchmark shortcuts. In this section, we showcase this aspect on the VSI-Super-Counting (VSC) benchmark. We (i) describe the VSC task, (ii) summarize the predictive sensing inference mechanism proposed in Cambrian-S, and (iii) show that a sanity check called \textit{VSC-Repeat}---repeating the same environments multiple times---drastically impacts performance. This reveals that the inference pipeline relies on a benchmark-specific shortcut of counting objects across new rooms rather than genuine spatial supersensing.\vspace{0.2cm}

\noindent\textbf{VSC Task.} VSC requires counting the number of \emph{unique} objects of a target category (e.g., chairs) across an entire long video stream. A correct solution must not only detect objects in individual frames, but also reason across time to identify duplicates and avoid counting the same object multiple times. The ground-truth label is a single unique-object count for the whole video. Performance is measured using mean relative accuracy~\citep{yang2025thinking}.\vspace{0.2cm}

\noindent\textbf{Inference mechanism of Cambrian-S for VSC.} Cambrian-S introduces a surprise-based event segmentation approach to improve performance on  VSC. It uses a latent frame-prediction model to compute a ``surprise'' signal (prediction error) and segments the continuous visual input whenever surprise is high, treating these points as scene-change boundaries. This yields a sequence of spatially coherent segments. Within each segment, the model accumulates frame features in an event buffer. When a high-surprise frame is detected, the model summarizes the buffered features to produce a segment-level count estimate, then clears the buffer and starts a new segment. This cycle repeats until the end of the video, after which the segment-level count estimates are aggregated into a final object-count prediction.  This design implicitly assumes that each segment corresponds to a distinct environment that is visited at most once. Under this assumption, resetting the buffer at each surprise boundary should not hurt unique-object counting (For details, see Section~4.3 in~\citep{cambrian-s}).\vspace{0.2cm}

\finding{}{\faBookmark~~The VSC-tailored inference method of Cambrian-S assumes that each segment corresponds to a distinct environment that is visited at most once.}\vspace{0.2cm}

\noindent\textbf{A Sanity Check.} We show that this bespoke segmentation mechanism described above inadvertently overfits to quirks of the VSC benchmark without truly improving spatial supersensing. For each 10-minute VSC video, we revisit the environments without introducing new objects by stitching the same clip to itself 1-5 times, forming our sanity check called \textit{VSC-Repeat}. This produces a sequence in which the same rooms and objects are revisited repeatedly, removing the shortcut in the VSC-benchmark of video never re-entering a previously visited environment. A model performing genuine spatial supersensing should recognize that these are repeated observations of the same environments and keep its predicted object count unchanged. VSC‑Repeat does not change the underlying spatial layout or object counts; it only violates the implicit assumption, built into both the benchmark and the predictive‑sensing inference pipeline, that each spatial environment will be seen once in a monotone sequence. \vspace{0.2cm}

\begin{figure}[!t]
    \centering
    \vspace{-0.4cm}
    \includegraphics[width=0.9\linewidth]{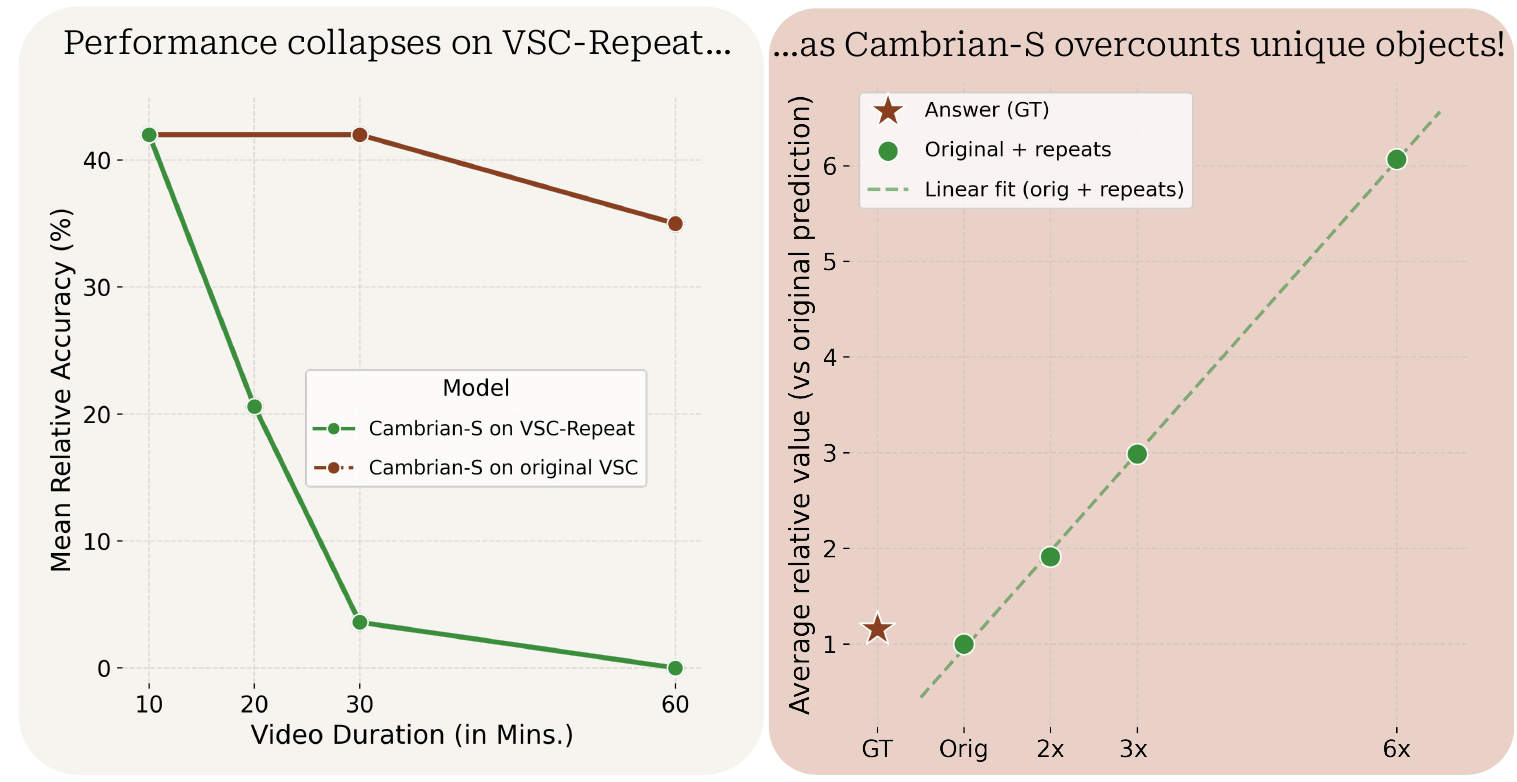}
    \vspace{-0.3cm}
        \caption{ \textbf{(Left) Cambrian collapses on \textit{VSC-Repeat}} with mean relative accuracy going from 42\% to 0\% after 5 repeats, indicating that its inference method relies on VSC-specific shortcuts rather than genuine spatial cognition. \textbf{(Right) Cambrian-S near-perfectly overcounts the objects proportional to repeats} indicating the predicted number of objects is strongly correlated to number of repeats, i.e. simply counting objects across new rooms as unique rather than maintaining a persistent set of unique objects across rooms. A supersensing model should output the same object counts across VSC-Repeat, since the unique object count stays the same. }
        \vspace{-0.3cm}
    \label{fig:vscrepeatedbench2}
\end{figure}

\noindent\textbf{Cambrian-S (almost) entirely fails on our VSC-Repeat Benchmark.} As illustrated in Figure~\ref{fig:vscrepeatedbench2}, repeating each 10-minute video 2 times (yielding a 30 min video) drastically reduces the mean relative accuracy (MRA) of the Cambrian-S inference pipeline from $42.0\%$ to $3.6\%$; the MRA further goes down to $0\%$ upon 5 repeats (yielding a 1 hour video).
By comparing performance of Cambrian-S on the original VSC benchmark (brown line), we show that the accuracy drop is not due to increasing the length of the video, but indeed due to our repeat procedure.
We also note that the predicted counts grow with the number of times the environments are revisited (Figure~\ref{fig:vscrepeatedbench2} right). This behavior is consistent with a shortcut heuristic that effectively counts \emph{segments} or \emph{rooms visited} rather than perform spatial supersensing and aggregate information throughout the video to recognize unique objects. This failure mode indicates that VSC performance is strongly tied to structural assumptions of the dataset (e.g., that scenes never repeat) rather than to revisit-invariant spatial reasoning. \vspace{0.2cm}

\finding{}{\faBookmark~~Cambrian-S' VSC-tailored inference strategy does not perform robust spatial supersensing or integrate information across repeated experiences. Instead, it exploits a quirk of the VSC benchmark: long videos composed of non-repeating environments.}\vspace{0.2cm}
 
\noindent \textbf{Conclusion.} A supersensing system that maintains an internal map over time should detect that revisited rooms do not introduce new objects; the current Cambrian‑S inference pipeline instead re‑counts at each surprise boundary. As a consequence, high VSC performance should be interpreted with caution when evaluating newer methods as evidence of continual counting, as they may exploit quirks of the VSC benchmark construction while neglecting the more advanced spatial and temporal understanding required for video world modeling. \vspace{0.2cm}

 \vspace{-0.2cm}
\section{Discussion}

\noindent Our findings do not imply that there has been no progress in improving spatial supersensing. Cambrian‑S reports that Gemini‑2.5‑Flash, used as a generic long‑context video model, achieves around 41.5\% accuracy on 60‑minute VSR videos -- similar in magnitude to Cambrian‑S model with predictive memory on the same split -- despite having no benchmark‑specific inference design exposed to users (see Figure 11 in~\citep{cambrian-s}). This illustrates that substantial VSR performance can arise from generic large‑scale video training and long context, without explicit supersensing‑targeted inference. Our main emphasis in this work is that (1) current spatial supersensing benchmarks do not sufficiently evaluate their spatial supersensing capabilities, and (2) current methods are specifically designed to target these benchmark idiosyncracies. For VSI-Super in particular, models can score well by encoding rules such as ``four object insertions'' or ``rooms never repeat'' rather than by building a stable internal map that survives revisits and perturbations. Looking forwards, we suggest some design principles that might partially fix these issues:\vspace{0.2cm}

\noindent \textbf{Invariance checks should be built in.} Benchmarks that claim to test long-horizon reasoning or spatial supersensing should include transformations that keep the ground truth fixed but alter superficial structure. Examples include repeating scenes, inserting revisits after long delays, shuffling segments with the same layout, or changing playback speed. Our proposed \textit{VSC-Repeat} shows how these checks reveal shortcuts: repeating the same video several times should leave the unique-object count unchanged, yet we find Cambrian-S' inference pipeline overcounts objects as if each segment were new. Designing such checks alongside benchmarks, and reporting performance under these perturbed conditions, makes it harder for methods to rely on benchmark-specific quirks.\vspace{0.2cm}

\noindent \textbf{Use more natural long-form video.} VSC aims to test accumulation of information and global coherence over long horizons, but its construction relies on “surprise” boundaries and curated episodes with non-repeating environments. This choice simplifies both data generation and inference, but it also encourages shortcuts that equate boundaries with environment changes. A more direct test of spatial supersensing would use truly long, continuous videos in which surprise does not reliably signal new environments -- e.g., egocentric traversals of buildings, streets, or synthetic worlds with natural revisits and loops \citep{singh2016krishnacam, Grauman_2024_CVPR}. In such settings, methods cannot safely treat every boundary as a new room and must track identity and layout explicitly.
\vspace{0.2cm}

\noindent \textbf{Emphasizing Open-Endedness and Independence of Modalities.} Benchmarks should stay aligned with realistic use cases. Benchmarks that collapse rich, open-ended tasks into narrow MCQs risk overstating practical capability and obscuring model limitations \citep{balepur2025these, chandak2025answer}. In multimodal understanding, most work has focused on inter-modality dependencies (how different modalities jointly relate to the output). However, information available in a single modality is often critical: We do not want self-driving cars to stop only when both video and audio agree that there is a pedestrian ahead. We believe more  work should model both intra-modality (the contribution of each modality on its own) and inter-modality dependencies (their joint relationships with the target task) to advance multimodal learning \citep{madaan2025multi, madaan2024jointly, zverev2025vggsounder}.\vspace{0.2cm}

\noindent \textbf{Meta-evaluation should be routine.} Our tools---\textit{NoSense} for VSR and \textit{VSC-Repeat} for VSC---act as simple stress tests of what benchmarks and methods actually require. We view this kind of meta-evaluation as analogous to ablation studies for models and believe it should be standard for benchmarks. For any proposed spatial supersensing task, we should ask: what is the simplest streaming baseline that exploits obvious regularities? Which invariances should hold, and do methods respect them? How does performance change when we break assumptions about segment length, revisit frequency, or segment structure? Answering these will help separate genuine advances in world modeling from improvements tied to a specific data generator.
Posing such questions has led to immense progress before in computer vision~\citep{beyer2020we}, language modelling~\citep{gema2025we,ghosh2025onebench,nezhurina2024alice}, compositionality~\citep{goodcrepe,hsieh2023sugarcrepe} and reasoning~\citep{hochlehnert2025sober,yao2025var}, and can hopefully further spur spatial supersensing as a more robust paradigm. 

\section{Conclusion}

In this work, we asked a simple question: \textit{do current spatial supersensing benchmarks and methods in fact require spatial supersensing?} For VSI-Super and Cambrian-S, the answer is largely \textit{no}. We uncover behaviors that illustrate benchmark–model co-adaptation. VSI-Super rewards bespoke inference recipes---one for recall and one for counting---that match the dataset’s generative assumptions. A model can score well by encoding rules such as “there are four object insertions” or “rooms never repeat”, rather than by learning capabilities that transfer when these assumptions change. On the other hand, the success of \textit{NoSense} shows the strength of contrastive vision–language representations as perception components of streaming systems. A simple SigLIP-based baseline can reproduce much of what current spatial supersensing methods achieve, with little machinery beyond pretrained vision and text encoders. Taken together, these findings indicate VSI-Super does not yet operationalize spatial supersensing: trivial admissible baselines can saturate key tasks without performing any long-horizon spatial integration. A genuine first step toward supersensing requires demonstrating at least one capability that trivial retrieval-based methods cannot achieve. Until then, we should view performance of future methods on VSI-Super with caution for claims about spatial supersensing or long-horizon world modeling.\vspace{0.2cm}

\noindent Going forward, we view \textit{world-model evaluation} as an important research problem, alongside world-modeling itself. Progress in spatial supersensing will require benchmarks that are robust to trivial transformations, explicitly require revisits to previous states and long-horizon integration, along with routinely stress-testing with adversarial baselines and perturbations. Our analyses move in this direction by exposing the fragility of current evaluations and by pointing to concrete, more faithful tests of video world models.
\section*{Limitations}

Our analysis has two main limitations. Empirically, we study a single benchmark family (VSI-Super) and one primary method (Cambrian-S). Other datasets or algorithms may behave differently, so our conclusions about spatial supersensing may not generalize. Conceptually, we diagnose failure modes and propose design principles, but we do not introduce a validated replacement benchmark. Building such benchmarks remains an open challenge and will require iterative cycles of task design, adversarial probing, and model development.\vspace{0.2cm}

\section*{Acknowledgments}

The authors would like to thank (in no particular order): Sebastian Dziadzio, Adhiraj Ghosh, Shashwat Goel, Matthias Kümmerer, Thao Nguyen, and Ronald Skorobogat for helpful feedback and comments on our draft. We thank the authors of Cambrian-S for engaging with our work and for their constructive feedback. This work is funded by the Federal Ministry of Education and Research (BMBF), FKZ: 01IS24079A. VU was supported by a Google PhD Fellowship in Machine Intelligence. VU also thanks the European Laboratory for Learning and Intelligent Systems (ELLIS) PhD program for support. AH and VU thank the International Max Planck Research School for Intelligent Systems (IMPRS-IS) for support. AH acknowledges funding by the Federal Ministry of Education and Research (BMBF), FKZ: 01IS24079A. AP and MB acknowledge acknowledges financial support by the Federal Ministry of Research, Technology and Space (BMFTR), FKZ: 16IS24085B and Open Philanthropy Foundation funded by the Good Ventures Foundation. This research utilized compute resources at the Tübingen Machine Learning Cloud, DFG FKZ INST 37/1057-1 FUGG.

{
    \small
    \bibliographystyle{plainnat}
    \bibliography{main}
}
\appendix
\clearpage

\section{Feedback from Cambrian-S Authors}
\label{sec:response}

We aim to foster open discussion in the spirit of journals such as \emph{Behavioral and Brain Sciences} (BBS), which use an open peer commentary format for controversial work. Before releasing this work, we emailed the Cambrian-S authors to share our findings. We present their response to give readers a more balanced perspective on this work:

\begin{tcolorbox}[colback=gray!10,colframe=black!50,title={Response from the Cambrian-S authors (November 20th, 2025)},]
\small{

Thank you very much for contacting us and sharing your findings! We truly appreciate the attention you have given to our work. Below, we have answered your questions, giving you our different views and some personal thoughts.\vspace{0.2cm}

After thoroughly reading your manuscript, we find your experimental settings to be sound. Some of your test setups are exactly what we had in mind! We appreciate your effort in designing this simple yet strong solution and sharing it with us.\vspace{0.2cm}

However, we would like to provide our different view as below:\vspace{0.2cm}

\textbf{First, we respectfully disagree the strong performance of NoSense implies VSR is ill-suited for supersensing.}\vspace{0.2cm}

A benchmark is always designed with specific purpose, to gauge specific capabilities or measure progress. However, once released, a benchmark can often be ``solved" in ways unintended by the design. The meaning of a result depends not only on the benchmark's design purpose but also on how it is used.\vspace{0.2cm}

VSR is like a ``Needle-In-A-Haystack" (NIAH) task for video. In the NLP world, NIAH tests a language model's ability to understand very long texts by hiding a specific ``needle" (like a random hash) in a long document. A trivial solution to solve this is to use a \textbf{search tool} (like CTRL+F). But NIAH is designed to \textbf{evaluate the LLM’s attention in long context}, not a truly hard problem in computer science or NLP. Most modern language models can easily solve this problem by writing code, however, we cannot claim that the resulting perfect performance in that way equates to perfect long-context understanding or the benchmark fails to achieve its purpose.\vspace{0.2cm}

Similarly, your solution perfectly solves any Video NIAH task. But this actually diverges from the intended purpose and expected usage of VSI-SUPER. Our goal is to build a smart MLLM that can handle long memories, not a specific tool that only works perfectly for ``video needle-in-the-haystack”.\vspace{0.2cm}

\textbf{Also, it is noteworthy that} we did think about this simple kind of solution when we designed VSR. That is why we assume the model \textbf{cannot know the search query while watching the video}. This is closer to how human memory and unconscious inference work: we notice and remember things without a specific question, and then we recall them only when a specific question is asked. Your email is a very helpful reminder that we did not explain this point clearly in our paper. \textbf{We will fix this and highlight it as soon as possible}.\vspace{0.2cm}

\textbf{Second}, about the VSC-Repeat failure, we know this is a limitation of our Cambrian-S (w/ Surprise Seg) model. While this might sound like an excuse, we actually chose not to include repeated scenes in the VSC design at first. Including repeats is more realistic but makes finding a strong and general solution much harder. Your email is a great reminder that we probably still need to add this setup to the next iteration of VSC.\vspace{0.2cm}

We truly appreciate your effort and for sharing your observations with us!\vspace{0.2cm}

As we stated in our paper, Cambrian-S and VSI-SUPER are just the \textbf{first step} toward \textbf{spatial supersensing}. Our current model and method are not yet ready for the complexities of the real world, and the benchmark is still limited compared to the full scope of spatial supersensing.\vspace{0.2cm}

We completely agree that we must do more to fix biases and make the benchmark more realistic. We are now working to gather more real-world, long video data (not just joined short clips) to build a better set of tests. We are also trying to improve our models to have better spatial sensing ability.\vspace{0.2cm}

\textbf{Finally, thank you again for pointing out the current limitations in both our benchmark and our model.}\vspace{0.2cm}

We'd be excited to see your findings published on arXiv, and we think it would be even stronger if you could include our response in the appendix. That way readers have the complete context.}
\end{tcolorbox}

\end{document}